\documentclass{article}

\usepackage{PRIMEarxiv}

\usepackage[utf8]{inputenc} 
\usepackage[T1]{fontenc}    
\usepackage{hyperref}       
\usepackage{url}            
\usepackage{booktabs}       
\usepackage{amsfonts}       
\usepackage{nicefrac}       
\usepackage{microtype}      
\usepackage{lipsum}
\usepackage{fancyhdr}       
\usepackage{graphicx}       
\graphicspath{{media/}}     
\usepackage{amsfonts}
\usepackage{amsmath}
\usepackage{bm}
\usepackage{amsthm}
\usepackage{amssymb}
\usepackage{graphicx}

\usepackage{multirow}
\usepackage[normalem]{ulem}
\useunder{\uline}{\ul}{}
\title{Continuous approximation by Convolutional Neural Networks with a sigmoidal function
}

\author{
  Weike Chang \\
  Nanchang University \\
  Nanchang\\
  \texttt{wkchang@foxmail.com} \\
}

\begin{document}
\maketitle

\begin{abstract}
In this paper we present a class of convolutional neural networks (CNNs) called non-overlapping CNNs in the study of approximation capabilities of CNNs. We prove that such networks with sigmoidal activation function are capable of approximating arbitrary continuous function defined on compact input sets with any desired degree of accuracy. This result extends existing results where only multilayer feedforward networks are a class of approximators. Evaluations elucidate the accuracy and efficiency of our result and indicate that the proposed non-overlapping CNNs are less sensitive to noise.
\end{abstract}


\section{Introduction}
The solution of inverse ill-posed problems has a widespread application in science and engineering, ranging from signal analysis~\cite{Nordebo2006} and communication systems~\cite{Ahmadzadeh2018} to medical instrumentation~\cite{Tatiana2019}, to astrophysics~\cite{Ramm1995} and high energy physics~\cite{Gainer2015}. We consider an inverse problem $Az=u$, with $A:F \rightarrow U$, $F$ and $U$ are metric spaces. The problem of determining the solution $z=A^{-1}\left(u\right)$ in the space $F$ from a set of initial data $u\in U$ is said to be stable on the metric spaces $F$ and $U$ if, for every positive number $\epsilon$, a positive number $\delta\left(\epsilon\right)$ exists, such that, if $d_U\left(u_1,u_2 \right)\leq\delta\left(\epsilon\right)$ then $d_F\left(z_1,z_2 \right)\leq\epsilon$, with $u_1,u_2\in U, z_1=R\left(u_1\right),z_2=R\left(u_2\right)$. In other words, the function $R$, which approximates  $A^{-1}$ and associates the initial data $u$ to the solution $z$, should be continuous, allowing a continuous dependence of the solution on a continuous change of the data.

With the rapid development of convolutional neural networks (CNNs), their capability  to solve the inverse problems has been investigated by many authors~\cite{Dong2016295,Kappeler2016109,Zhang2019883,TIAN2020461,Huang2020324}. While the class of functions represented with CNNs exhibit stability in the solution of specific inverse problems~\cite{Jin20174509}, their generalization is far to be achieved. In other words, the stability obtained by CNN may be just applied to specific experiments, but cannot be generalized to any class of inverse problems, due to a lack of theoretical fundament. Here the key-problem is to understand whether CNNs approximate any class of continuous functions, or they restrict the solution to a particular group of functions.

In other words, an essential condition for the applicability of CNNs to the solution of ill-posed problems is that the set of maps $G$ represented by CNNs are dense in the set of all continuous functions $F$ representing the solution of the inverse problem. Therefore for every positive real number $\varepsilon$ and for every $f \in F$ there exists $g \in G$ such that $\left\| {f - g} \right\| < \varepsilon$. In this paper, $\left\|  \cdot  \right\|$ is strictly defined by the metric $\left\| {f - g} \right\| = \sup \left\{ {\left| {f\left( x \right) - g\left( x \right)} \right|,x \in X} \right\}$. If we assume $g$ can approximate $f$, then there exists $\varepsilon >0$ such that $\left\| {g\left( {{x_1}} \right) - g\left( {{x_2}} \right)} \right\| \le \left\| {g\left( {{x_1}} \right) - f\left( {{x_1}} \right)} \right\| + \left\| {f\left( {{x_1}} \right) - f\left( {{x_2}} \right)} \right\| + \left\| {f\left( {{x_2}} \right) - g\left( {{x_2}} \right)} \right\| < \varepsilon $ for every $x_1,x_2 \in X$ with $\left\| {{x_1} - {x_2}} \right\| < \delta$.

Hornik et al \cite{Hornik1989359}, Hornik \cite{Hornik1991251} and Cybenko \cite{Cybenko1989303} gave a direct proof of the approximation capabilities of standard multilayer feedforward networks with a single hidden layer. K$\mathring{\rm u}$rkov\'{a} \cite{KURKOVA1992501} took full advantage of Kolmogorov's representation theorem to prove that any continuous functions defined on a compact set can be uniformly learned by means of perceptron type networks with two hidden layers and sigmoidal activation functions (i.e., a function $\varphi $ with $\mathop {\lim }\limits_{x \to + \infty } \varphi \left( x \right) = 1$ and $\mathop {\lim }\limits_{x \to  - \infty } \varphi \left( x \right) = 0$). However, such results are only appropriate for networks where the neurons in each hidden layer are fully connected to outputs of the previous layer.

Non-overlapping CNNs overcome these limitations. They are defined as a class of CNNs where the length of the output in each convolutional layer can be obtained by dividing the length of the input by the length of the convolution kernel, since the length of convolution kernels equals the length of stride. In this paper, extending the results from \cite{Hornik1989359}, we demonstrate the following theorems.

\newtheorem{theorem}{Theorem}
\begin{theorem}\label{theorem1}
Let $n,k \in \mathbb{N}\equiv \left\{ {1,2, \ldots } \right\}$ with ${n \mathord{\left/
 {\vphantom {n k}} \right.
 \kern-\nulldelimiterspace} k} \in \mathbb{N}$, $\textbf{{x}} = \left( {{x_0},{x_1}, \cdots ,{x_{n - 1}}} \right)$ belong to a compact set $ E^n$, $k$ be the length of convolution kernel $\textbf{{w}}$ from the input $\textbf{{x}}$ to the hidden layer and 1 be the length of convolution kernel $\bm{\beta}$ from the hidden layer to the output layer. Then the non-overlapping CNN $\textbf{{g}}$ with a single hidden layer and a sigmoidal function $\varphi $, can approximate an arbitrary continuous function defined on $E^n$, provided only that sufficiently many $\textbf{{w}}$ are available.
 \end{theorem}

 Theorem \ref{theorem1} improves the applicability of the result in \cite{Hornik1989359} to show that non-overlapping CNNs with a single hidden layer and sigmoidal activation function can approximate arbitrary continuous functions on a compact subset of an arbitrarily finite dimensional space. The proof is left to Section \ref{section2}. The limitation of this theorem is that the length of the output networks must be equal to ${{n \mathord{\left/
 {\vphantom {n k}} \right.
 \kern-\nulldelimiterspace} k}}$. We therefore extend it as follows:

\newtheorem{theorem1}[theorem]{Theorem}
\begin{theorem1}\label{theorem2}
 Let $\textbf{{w}}$, $\textbf{{x}}$ and $\textbf{{g}}$ be as defined in Theorem \ref{theorem1}, $M,n \in \mathbb{N}$, ${{k_i}} \in \mathbb{N}$ with $i = 1,2, \cdots ,M$ be the length of convolution kernels $\textbf{{w}}_i$ in network $\textbf{{g}}_i$ and $n = \prod\nolimits_{i = 1}^M {{k_i}} $. Then the non-overlapping CNN $G$ with $M$ networks $\textbf{{g}}_i$ are capable of approximating an arbitrary continuous real-valued function on $E^n$, provided only that sufficiently many $\textbf{{w}}_i$ are available.
 \end{theorem1}

 Theorem \ref{theorem2} overcomes the limitation from Theorem \ref{theorem1} to show that non-overlapping CNNs with one output neuron using sigmoidal activation functions are capable of approximating arbitrary continuous real-valued functions defined on a compact subset of an arbitrarily finite dimensional space. The proof is left to Section \ref{section3} where we convert the theoretical proof of approximation capability of the above network $G$ to the problem of approximating a continuous composite function with a composite mapping. Theorem \ref{theorem2} establishes therefore that the class of functions represented by non-overlapping CNNs can be used to represent the approximated solution of any class of ill-posed inverse problems.

\section{Proof of Theorem \ref{theorem1}}
\label{section2}
Since values of possible inputs are bounded in practical applications, we shall consider that they are within a compact subset ${E^n}$ of a $n$-dimensional space ${\mathbb{R}}^n$ (where $\mathbb{R}$ denotes the set of real numbers). Moreover, we denote the set of all continuous functions defined on an arbitrarily finite dimensional space $X$ by $C^0 \left( X \right)$. Here, we introduce the following definition which allows us to precisely understand the operand calculation of the convolutional layer containing a single convolution kernel under consideration.
\newtheorem{definition}[theorem]{Definition}
 \begin{definition}\label{definition1}
 For $n,k \in \mathbb{N}$ with ${n \mathord{\left/
 {\vphantom {n k}} \right.
 \kern-\nulldelimiterspace} k} \in \mathbb{N}$, $\textbf{{x}}_j = \left( {x_{\left( {j - 1} \right)k},{x_{\left( {j - 1} \right)k+1}}, \cdots ,{x_{jk - 1}}} \right) $ with $j = 1, \cdots ,{n \mathord{\left/{\vphantom {n k}} \right.\kern-\nulldelimiterspace} k}$ and $\textbf{{x}}=\left( {{\textbf{{x}}_1},{\textbf{{x}}_2}, \cdots ,{\textbf{{x}}_{{n \mathord{\left/
 {\vphantom {n k}} \right.\kern-\nulldelimiterspace} k}}}} \right) \in E^n$. $A^k$ is the set of all affine functions from ${E^k}$ to $\mathbb{R}$, that is, the set of all functions of the form ${A\left( \textbf{{x}}_j \right)} = \textbf{{w}} \cdot {\textbf{{x}}_j} + b = \sum\nolimits_{i = \left( {j - 1} \right)k}^{jk - 1} {{w_{i - \left( {j - 1} \right)k}}} {x_i} + b$ where $\textbf{{w}} \in {\mathbb{R}^k}$ and $b \in \mathbb{R}$. Let $\left(\textbf{{A}}\textbf{{x}}\right)=\left( {A\left( {{\textbf{{x}}_1}} \right),A\left( {{\textbf{{x}}_2}} \right), \cdots , A\left({{\textbf{{x}}_{{n \mathord{\left/{\vphantom {n k}} \right. \kern-\nulldelimiterspace} k}}}} \right)} \right)^T$ is a vector in ${\mathbb{R}^{{n \mathord{\left/
 {\vphantom {n k}} \right.\kern-\nulldelimiterspace} k}}}$.
 \end{definition}

In this context, $\textbf{\emph{x}}$ corresponds to the input, $\textbf{\emph{A}}$ corresponds to the output, $\textbf{\emph{x}}_j$ corresponds to $jth$ part of $\textbf{\emph{x}}$, $b$ corresponds to a bias and $\textbf{\emph{w}}$ corresponds to a convolution kernel with $k$ weights.

In the middle layer of CNNs, a convolutional layer usually contains many convolution kernels and is followed by a activation layer. These two layers are combined into a new layer called hidden layer, and the convolution kernel $\textbf{\emph{w}}$ is also called the hidden neuron. We formally define hidden layer as follows.

\newtheorem{definition2}[theorem]{Definition}
\begin{definition2}\label{definition2}
For $q \in \mathbb{N}$, $A_i \in  A^k$ with $i=1,2,\cdots,q$ and an activation function $\varphi $ mapping $\mathbb{R}$ to $\mathbb{R}$. $\textbf{{M}}\left( \textbf{{x}}\right) =\left( \varphi \left( \textbf{{A}}_1 \left( \textbf{{x}} \right) \right), \varphi \left( \textbf{{A}}_2 \left( \textbf{{x}} \right) \right), \cdots, \varphi \left( \textbf{{A}}_q \left( \textbf{{x}} \right) \right) \right)$ is the output function of the hidden layer with $q$ convolution kernels, where $\textbf{{A}}_i \left( \textbf{{x}} \right) =\left( A_i\left( {{\textbf{{x}}_1}} \right),A_i\left( {{\textbf{{x}}_2}} \right), \cdots ,A_i\left( {{\textbf{{x}}_{{n} \mathord{\left/{\vphantom {{n} k}} \right.\kern-\nulldelimiterspace} k}}}\right) \right)^T$.
\end{definition2}

We present the following lemma from \cite{Hornik1989359} to express a pleasant property of single hidden layer feedforward networks with sigmoidal activation function.

\newtheorem{lemma}[theorem]{Lemma}
\begin{lemma}\label{lemma1}
Let $d, q \in \mathbb{N}$, $E^d$ be a compact set, and $\varphi $ be a sigmoidal function. Then the set of all functions $\sum g$ of the form $g\left( \textbf{{x}} \right) = \sum\nolimits_{i = 1}^q {{\beta _i}} \varphi \left( {\sum\nolimits_{j = 0}^{d - 1} {w_j^i{x_j}} + {b_i}} \right)$ \\, where $\textbf{{x}} \in {E^d}$ and ${\beta _i},{b_i},{w_j^i} \in \mathbb{R}$, is uniformly dense in ${C^0}\left( {{E^d}} \right)$.
\end{lemma}

In the context of neural networks, Lemma \ref{lemma1} expresses that standard multilayer feedforward networks with a single hidden layer and a sigmoidal activation function are capable of approximating arbitrary continuous functions on a compact set with an arbitrary accuracy when $q$ is sufficiently large.

First of all, we study the continuity of a function that maps arbitrary finite dimensional space to another space.

\newtheorem{theorem2}[theorem]{Theorem}
\begin{theorem2}\label{theorem3}
 Let $\textbf{{x}}$ and $\textbf{{x}}_j$ be as defined in Definition \emph{\ref{definition1}}, $n,k \in \mathbb{N}$ with ${n \mathord{\left/ {\vphantom {n k}} \right.
\kern-\nulldelimiterspace} k} \in \mathbb{N}$ and $f$ be a continuous function from ${E^k}$ to $\mathbb{R}$. Then $\textbf{{f}}\left( \textbf{{x}} \right) = \left( f\left( {{\textbf{{x}}_1}} \right), f\left( {{\textbf{{x}}_2}} \right), \cdots , \right. \\ \left. f\left( {{\textbf{{x}}_{{n \mathord{\left/{\vphantom {n k}} \right.
 \kern-\nulldelimiterspace} k}}}} \right) \right)$ is a continuous function from $E^n$ to a compact subset of ${\mathbb{R}^{{n \mathord{\left/{\vphantom {n k}} \right. \kern-\nulldelimiterspace} k}}}$.
\end{theorem2}
\begin{proof}[\rm \textbf{Proof.}]
Let $\textbf{\emph{z}}=\left( {{\textbf{\emph{z}}_1},{\textbf{\emph{z}}_2}, \cdots ,{\textbf{\emph{z}}_{{n \mathord{\left/
 {\vphantom {n k}} \right.
 \kern-\nulldelimiterspace} k}}}} \right) \in {E^n}$, then there exists $\delta  > 0$ such that $\left\| {{\textbf{\emph{x}}_j} - {\textbf{\emph{z}}_j}} \right\| < \delta $ for $\textbf{\emph{x}},\textbf{\emph{z}}\in {E^n}$ with $\left\| {{\textbf{\emph{x}}} - {\textbf{\emph{z}}}} \right\| < \delta $. And since $f$ is a continuous function from $E^k$ to $\mathbb{R}$, there exists $\varepsilon >0 $ such that $\left\| {f\left( {{\textbf{\emph{x}}_j}} \right) - f\left( {{\textbf{\emph{z}}_j}} \right)} \right\| < {{\varepsilon k} \mathord{\left/ {\vphantom {{\varepsilon k} n}} \right. \kern-\nulldelimiterspace} n}$ for $\left\| {{\textbf{\emph{x}}_j} - {\textbf{\emph{z}}_j}} \right\| < \delta $. Then
\begin{equation}\label{eq1}
\left\| {\textbf{\emph{f}}\left( \textbf{\emph{x}} \right) - \textbf{\emph{f}}\left( \textbf{\emph{z}} \right)} \right\| \le \sum\nolimits_{j = 1}^{\frac{n}{k}} {\left\| {f\left( {{\textbf{\emph{x}}_j}} \right) - f\left( {{\textbf{\emph{z}}_j}} \right)} \right\|} < \frac{n}{k} \times \frac{{k\varepsilon }}{n} = \varepsilon
\end{equation}
for $\left\| {{\textbf{\emph{x}}} - {\textbf{\emph{z}}}} \right\| < \delta$. Hence, $\textbf{\emph{f}}{\rm{:}}{E^n} \to {\mathbb{R}^{{n \mathord{\left/ {\vphantom {n k}} \right. \kern-\nulldelimiterspace} k}}}$ is a continuous function, and $\textbf{\emph{f}}\left( {{E^n}} \right)$ is a compact set since $E^n$ is a compact set.
\end{proof}

Using Lemma \ref{lemma1} and Theorem \ref{theorem3}, we now are able to prove Theorem \ref{theorem1}:
\begin{proof}[\rm \textbf{Proof.}]
Since $\textbf{\emph{g}}$ is a non-overlapping CNN and $k$ is the length of the convolution kernel $\textbf{\emph{w}}$, the length of each feature vector of the hidden layer is ${{n \mathord{\left/{\vphantom {n k}} \right.\kern-\nulldelimiterspace} k}}$. Let $q \in \mathbb{N}$ be the number of hidden neurons $\textbf{\emph{w}}$, then derived from Definition \ref{definition2}, we can write the output matrix $\textbf{\emph{M}}$ of the hidden layer as follows:
\begin{equation*}
 \textbf{\emph{M}} = \left( {\begin{array}{*{20}{c}}
{\varphi \left( {\sum\nolimits_{j = 0}^{k - 1} {w_j^1{x_j} + {b_1}} } \right)}& \cdots &{\varphi \left( {\sum\nolimits_{j = 0}^{k - 1} {w_j^q{x_j} + {b_q}} } \right)}\\
{\begin{array}{*{20}{c}}
{\varphi \left( {\sum\nolimits_{j = k}^{2k - 1} {w_{j - k}^1{x_j} + {b_1}} } \right)}\\
 \vdots
\end{array}}&{\begin{array}{*{20}{c}}
 \cdots \\
 \ddots
\end{array}}&{\begin{array}{*{20}{c}}
{\varphi \left( {\sum\nolimits_{j = k}^{2k - 1} {w_{j - k}^q{x_j} + {b_q}} } \right)}\\
 \vdots
\end{array}}\\
{\varphi \left( {\sum\nolimits_{j = n - k}^{n - 1} {w_{j - n + k}^1{x_j} + {b_1}} } \right)}& \ldots &{\varphi \left( {\sum\nolimits_{j = n - k}^{n - 1} {w_{j - n + k}^q{x_j} + {b_q}} } \right)}
\end{array}} \right)
\end{equation*}
where $\textbf{\emph{M}}$ is a $\left( {{n \mathord{\left/{\vphantom {n k}} \right.\kern-\nulldelimiterspace} k}} \right) \times q$ matrix and each column of $\textbf{\emph{M}}$ represents a feature vector, $b$ represents the bias term. Since the number of $\bm{\beta}$ is 1, that is, $\bm{\beta}  = \left( {{\beta _1},{\beta _2}, \cdots {\beta _q}} \right)$, the network output $\textbf{\emph{g}}\left( \textbf{\emph{x}} \right)$ is written as follows: \begin{equation*}
\textbf{\emph{g}}\left( \textbf{\emph{x}} \right) = \textbf{\emph{M}}{\bm{\beta}^T} = \left( {\begin{array}{*{20}{c}}
{\sum\nolimits_{i = 1}^q {{\beta _i}\varphi \left( {{\textbf{\emph{w}}^i} \cdot {\textbf{\emph{x}}_1} + {b_i}} \right)} }\\
{\sum\nolimits_{i = 1}^q {{\beta _i}\varphi \left( {{\textbf{\emph{w}}^i} \cdot {\textbf{\emph{x}}_2} + {b_i}} \right)} }\\
 \vdots \\
{\sum\nolimits_{i = 1}^q {{\beta _i}\varphi \left( {{\textbf{\emph{w}}^i} \cdot {\textbf{\emph{x}}_\frac{n}{k}} + {b_i}} \right)} }
\end{array}} \right) = \left( {\begin{array}{*{20}{c}}
{\sum\nolimits_{i = 1}^q {{\beta _i}\varphi \left( {{A_i}\left( {{\textbf{\emph{x}}_1}} \right)} \right)} }\\
{\sum\nolimits_{i = 1}^q {{\beta _i}\varphi \left( {{A_i}\left( {{\textbf{\emph{x}}_2}} \right)} \right)} }\\
 \vdots \\
{\sum\nolimits_{i = 1}^q {{\beta _i}\varphi \left( {{A_i}\left( {{\textbf{\emph{x}}_{\frac{n}{k}}}} \right)} \right)} }
\end{array}} \right)
\end{equation*}
where $\textbf{\emph{x}}_j=\left( {{x_{\left( {j - 1} \right)k}},{x_{\left( {j - 1} \right)k + 1}}, \cdots ,{x_{jk - 1}}} \right) \in E^k$ with $j= 1, \cdots ,{n \mathord{\left/
{\vphantom {n k}} \right.\kern-\nulldelimiterspace} k}$ and ${\textbf{\emph{w}}^i}  = \left( {w_0^i,w_1^i, \cdots ,w_{k - 1}^i} \right) \in \mathbb{R}^k$. We can observe that each element of $\textbf{\emph{g}}\left( \textbf{\emph{x}} \right)$ belongs to $\sum g $ from Lemma \ref{lemma1}. So we have $\textbf{\emph{g}}\left( \textbf{\emph{x}} \right)=\left( {g\left( {{\textbf{\emph{x}}_1}} \right),g\left( {{\textbf{\emph{x}}_2}} \right), \cdots ,g\left( {{\textbf{\emph{x}}_{{n \mathord{\left/
{\vphantom {n k}} \right.\kern-\nulldelimiterspace} k}}}} \right)} \right)^T$. We introduce Figure \ref{fig:1} which allows us to better understand $\textbf{\emph{g}}\left( \textbf{\emph{x}} \right)$. Let $f$ be a continuous function from ${E^k}$ to $\mathbb{R}$, then following Lemma \ref{lemma1}, there exists $\varepsilon > 0$ such that $\left\| {g\left( {{\textbf{\emph{x}}_j}} \right) - f\left( {{\textbf{\emph{x}}_j}} \right)} \right\|{{ < k\varepsilon } \mathord{\left/
{\vphantom {{ < k\varepsilon } n}} \right.\kern-\nulldelimiterspace} n}$. And following Theorem \ref{theorem3}, it is easy to show that $\textbf{\emph{f}}\left( \textbf{\emph{x}} \right) = \left( {f\left( {{\textbf{\emph{x}}_1}} \right), \cdots ,f\left( {{\textbf{\emph{x}}_{{n \mathord{\left/
{\vphantom {n k}} \right.\kern-\nulldelimiterspace} k}}}} \right)} \right)^T$ is a continuous function from $E^n$ to a compact subset of ${\mathbb{R}^{{n \mathord{\left/{\vphantom {n k}} \right.\kern-\nulldelimiterspace} k}}}$. Then \begin{equation}\label{eq2}
\left\| {\textbf{\emph{g}}\left( \textbf{\emph{x}} \right) - \textbf{\emph{f}}\left( \textbf{\emph{x}} \right)} \right\| \le \sum\nolimits_{j = 1}^{\frac{n}{k}} {\left\| {g\left( {{\textbf{\emph{x}}_j}} \right) - f\left( {{\textbf{\emph{x}}_j}} \right)} \right\|} \\
 < \frac{n}{k} \times \frac{{k\varepsilon }}{n} = \varepsilon
\end{equation}
and $\textbf{\emph{g}}\left( E^n \right)$ is bounded. This completes the proof.
\end{proof}

In other words, non-overlapping CNNs with a single hidden layer followed by a unbiased convolutional layer are capable of arbitrarily accurate approximation to any continuous functions from a compact subset of $n$-dimensional space to ${{{n} \mathord{\left/ {\vphantom { {n} k}} \right. \kern-\nulldelimiterspace} k}}$-dimensional space when feeding $\textbf{\emph{x}} \in E^n$ into such networks. It is worth noting that Theorem \ref{theorem1} is the same as Lemma \ref{lemma1} when $ k=n$.

\section{Proof of Theorem \ref{theorem2}}
\label{section3}
In this paper, non-overlapping CNNs under consideration are a class of multilayer networks. Hence, the output function of such a network is represented by the composition of output functions of the previous layers, that is, it is a composite function. In what follows, the composite function $\textbf{\emph{f}}\left( \textbf{\emph{x}} \right)=\textbf{\emph{f}}_2\left( \textbf{\emph{f}}_1\left( \textbf{\emph{x}} \right) \right)$ is abbreviated as $\textbf{\emph{f}}\left( \textbf{\emph{x}} \right)=\textbf{\emph{f}}_2 \ensuremath\mathop{\scalebox{.65}{$\circ$}} \textbf{\emph{f}}_1\left( \textbf{\emph{x}} \right)$, where $\textbf{\emph{f}}_1$ and $\textbf{\emph{f}}_2$ are internal functions of $\textbf{\emph{f}}$. We note that composite functions have a pleasant property when internal functions of them are continuous functions. This leads to the following lemma.

\newtheorem{lemma1}[theorem]{Lemma}
\begin{lemma1}\label{lemma2}
Let $M \in \mathbb{N} $, $\textbf{{f}}_1$ be continuous in metric space $X$ and $\textbf{{f}}_i$, where $i= 2,3,\cdots,M$, continuous in ${\textbf{{f}}_{i - 1}} \ensuremath\mathop{\scalebox{.65}{$\circ$}} {\textbf{{f}}_{i-2}} \ensuremath\mathop{\scalebox{.65}{$\circ$}} \cdots \ensuremath\mathop{\scalebox{.65}{$\circ$}}{\textbf{{f}}_1}$. Then $\textbf{{F}}=\textbf{{f}}_{M}\ensuremath\mathop{\scalebox{.65}{$\circ$}} \textbf{{f}}_{M-1} \ensuremath\mathop{\scalebox{.65}{$\circ$}} \cdots \ensuremath\mathop{\scalebox{.65}{$\circ$}}\textbf{{f}}_1$ is continuous in $X$.
\end{lemma1}
\begin{proof}[\rm \textbf{Proof.}]
If we assume $M=2$, then $\textbf{\emph{F}}=\textbf{\emph{f}}_{2} \ensuremath\mathop{\scalebox{.65}{$\circ$}} {\textbf{\emph{f}}_{1}}$. Since $\textbf{\emph{f}}_1$ is continuous in $X$, there exists $\delta ,\eta  > 0$ such that $\left\| {\textbf{\emph{f}}_1\left( \textbf{\emph{x}} \right) - \textbf{\emph{f}}_1\left( \textbf{\emph{x}}^* \right)} \right\|< \eta $ for every $\textbf{\emph{x}},\textbf{\emph{x}}^* \in X$ with $\left\| {\textbf{\emph{x}} - \textbf{\emph{x}}^*} \right\| < \delta $. And since $\textbf{\emph{f}}_2$ is continuous in $\textbf{\emph{f}}_1 \left( X \right)$, there must exist $\kappa > 0$ such that $\left\| \textbf{\emph{F}}\left( \textbf{\emph{x}} \right) - \textbf{\emph{F}}\left( \textbf{\emph{x}}^* \right) \right\|=\left\| \textbf{\emph{f}}_2\ensuremath\mathop{\scalebox{.65}{$\circ$}}  {\textbf{\emph{f}}_1\left( \textbf{\emph{x}} \right)} - \textbf{\emph{f}}_2\ensuremath\mathop{\scalebox{.65}{$\circ$}}  {\textbf{\emph{f}}_1\left( \textbf{\emph{x}}^* \right)} \right\| < \kappa$. Therefore, $\textbf{\emph{F}}=\textbf{\emph{f}}_{2} \ensuremath\mathop{\scalebox{.65}{$\circ$}} {\textbf{\emph{f}}_{1}}$ is continuous in $X$. If we assume $M=3$, then $\textbf{\emph{F}}=\textbf{\emph{f}}_{3} \ensuremath\mathop{\scalebox{.65}{$\circ$}} \textbf{\emph{f}}_{2}\ensuremath\mathop{\scalebox{.65}{$\circ$}} \textbf{\emph{f}}_{1}$. Let $\textbf{\emph{v}}=\textbf{\emph{f}}_{2} \ensuremath\mathop{\scalebox{.65}{$\circ$}} \textbf{\emph{f}}_{1}$, then from the above, $\left\| \textbf{\emph{v}}\left( \textbf{\emph{x}} \right) - \textbf{\emph{v}}\left( \textbf{\emph{x}}^* \right) \right\| < \kappa$ for $\left\| {\textbf{\emph{x}} - \textbf{\emph{x}}^*} \right\| < \delta$. Since $\textbf{\emph{f}}_3$ is continuous in $\textbf{\emph{v}}\left( X \right)$, there exists $\varepsilon > 0$ such that $\left\| \textbf{\emph{F}}\left( \textbf{\emph{x}} \right) - \textbf{\emph{F}}\left( \textbf{\emph{x}}^* \right) \right\|=\left\| \textbf{\emph{f}}_3\ensuremath\mathop{\scalebox{.65}{$\circ$}} {\textbf{\emph{v}}\left( \textbf{\emph{x}} \right)}  - \textbf{\emph{f}}_3\ensuremath\mathop{\scalebox{.65}{$\circ$}}  {\textbf{\emph{v}}\left( \textbf{\emph{x}}^* \right)} \right\| < \varepsilon$. In other words, $\textbf{\emph{F}}=\textbf{\emph{f}}_{3} \ensuremath\mathop{\scalebox{.65}{$\circ$}} \textbf{\emph{f}}_{2}\ensuremath\mathop{\scalebox{.65}{$\circ$}} \textbf{\emph{f}}_{1}$ is continuous in $X$. By following the line of proof of the above, we can finally prove that $\textbf{\emph{F}}=\textbf{\emph{f}}_{M}\ensuremath\mathop{\scalebox{.65}{$\circ$}} \textbf{\emph{f}}_{M-1} \ensuremath\mathop{\scalebox{.65}{$\circ$}} \cdots \ensuremath\mathop{\scalebox{.65}{$\circ$}}\textbf{\emph{f}}_1$ is also continuous in $X$.
\end{proof}

On the basis of the results in the previous section and Lemma \ref{lemma2}, it is possible to prove the validity of Theorem \ref{theorem2} as follows:
\begin{proof}[\rm \textbf{Proof.}]
Since the non-overlapping CNN $G$ is a cascade network and consists of $M$ networks $\textbf{\emph{g}}_i$, the formula of $G$ can be written as $G=g_{M}\ensuremath\mathop{\scalebox{.65}{$\circ$}} \cdots \ensuremath\mathop{\scalebox{.65}{$\circ$}} \textbf{\emph{g}}_2\ensuremath\mathop{\scalebox{.65}{$\circ$}}\textbf{\emph{g}}_1$, where network $\textbf{\emph{g}}_1$ is a mapping from $E^n$ to ${\mathbb{R}^{{n \mathord{\left/
{\vphantom {n {{k_1}}}} \right. \kern-\nulldelimiterspace} {{k_1}}}}}$, network $\textbf{\emph{g}}_i$ with $i=2,3, \cdots ,M - 1$ is a mapping from ${\mathbb{R}^{{n \mathord{\left/
{\vphantom {n {\prod\nolimits_{j = 1}^{i - 1} {{k_j}} }}} \right. \kern-\nulldelimiterspace} {\prod\nolimits_{j = 1}^{i - 1} {{k_j}} }}}}$ to ${\mathbb{R}^{{n \mathord{\left/
{\vphantom {n {\prod\nolimits_{j = 1}^i {{k_j}} }}} \right.\kern-\nulldelimiterspace} {\prod\nolimits_{j = 1}^i {{k_j}} }}}}$, and network $g_M$ is a mapping from ${\mathbb{R}^{{n \mathord{\left/ {\vphantom {n {\prod\nolimits_{j = 1}^{M - 1} {{k_j}} }}} \right.\kern-\nulldelimiterspace} {\prod\nolimits_{j = 1}^{M - 1} {{k_j}} }}}}$ to $\mathbb{R}$ because $n = \prod\nolimits_{i = 1}^M {{k_i}} $. Let $\textbf{\emph{f}}_1$, $\textbf{\emph{f}}_i$ with $i=2,3, \cdots ,M - 1$ and $f_M$ be continuous mappings from $E^n$ to ${\mathbb{R}^{{n \mathord{\left/{\vphantom {n {{k_1}}}} \right. \kern-\nulldelimiterspace} {{k_1}}}}}$, ${R^{{n \mathord{\left/
{\vphantom {n {\prod\nolimits_{j = 1}^{i - 1} {{k_j}} }}} \right. \kern-\nulldelimiterspace} {\prod\nolimits_{j = 1}^{i - 1} {{k_j}} }}}}$ to ${\mathbb{R}^{{n \mathord{\left/
{\vphantom {n {\prod\nolimits_{j = 1}^i {{k_j}} }}} \right. \kern-\nulldelimiterspace} {\prod\nolimits_{j = 1}^i {{k_j}} }}}}$ and ${\mathbb{R}^{{n \mathord{\left/
{\vphantom {n {\prod\nolimits_{j = 1}^{M - 1} {{k_j}} }}} \right. \kern-\nulldelimiterspace} {\prod\nolimits_{j = 1}^{M - 1} {{k_j}} }}}}$ to $\mathbb{R}$, respectively. Then according to Lemma \ref{lemma2}, $F=f_{M}\ensuremath\mathop{\scalebox{.65}{$\circ$}} \cdots \ensuremath\mathop{\scalebox{.65}{$\circ$}} \textbf{\emph{f}}_2\ensuremath\mathop{\scalebox{.65}{$\circ$}}\textbf{\emph{f}}_1$ is a continuous function from $E^n$ to $\mathbb{R}$. If we assume $M=2$, then
\begin{equation}\begin{aligned}
\left\| {G \left(\textbf{\emph{x}}\right) - F \left(\textbf{\emph{x}}\right)} \right\|&=\left\|g_2\ensuremath\mathop{\scalebox{.65}{$\circ$}}\textbf{\emph{g}}_1\left(\textbf{\emph{x}}\right)- f_2\ensuremath\mathop{\scalebox{.65}{$\circ$}}\textbf{\emph{f}}_1\left(\textbf{\emph{x}}\right) \right\| \\
&=\left\| g_2\ensuremath\mathop{\scalebox{.65}{$\circ$}}\textbf{\emph{g}}_1\left(\textbf{\emph{x}}\right)- f_2\ensuremath\mathop{\scalebox{.65}{$\circ$}}\textbf{\emph{g}}_1\left(\textbf{\emph{x}}\right)+ f_2\ensuremath\mathop{\scalebox{.65}{$\circ$}}\textbf{\emph{g}}_1\left(\textbf{\emph{x}}\right)- f_2\ensuremath\mathop{\scalebox{.65}{$\circ$}}\textbf{\emph{f}}_1\left(\textbf{\emph{x}}\right) \right\| \\
&\le \left\| g_2\ensuremath\mathop{\scalebox{.65}{$\circ$}}\textbf{\emph{g}}_1\left(\textbf{\emph{x}}\right)- f_2\ensuremath\mathop{\scalebox{.65}{$\circ$}}\textbf{\emph{g}}_1\left(\textbf{\emph{x}}\right)\right\|+ \left\|f_2\ensuremath\mathop{\scalebox{.65}{$\circ$}}\textbf{\emph{g}}_1\left(\textbf{\emph{x}}\right)- f_2\ensuremath\mathop{\scalebox{.65}{$\circ$}}\textbf{\emph{f}}_1\left(\textbf{\emph{x}}\right) \right\|
\end{aligned}
\label{eq3}
\end{equation}
For every $\textbf{\emph{x}} \in E^n$, there exists $\tau >0$ such that $\left\|\textbf{\emph{g}}_1\left(\textbf{\emph{x}}\right)-\textbf{\emph{f}}_1\left(\textbf{\emph{x}}\right)\right\| <\tau$ according to Theorem \ref{theorem1}. Since $f_2$ is continuous in ${\mathbb{R}^{{n \mathord{\left/{\vphantom {n {{k_1}}}} \right.
\kern-\nulldelimiterspace} {{k_1}}}}}$, there exists $\delta> 0$ such that $\left\|f_2 \ensuremath\mathop{\scalebox{.65}{$\circ$}}\textbf{\emph{g}}_1\left(\textbf{\emph{x}}\right)- f_2\ensuremath\mathop{\scalebox{.65}{$\circ$}}\textbf{\emph{f}}_1\left(\textbf{\emph{x}}\right)\right\|<{\delta  \mathord{\left/ {\vphantom {\delta  4}} \right. \kern-\nulldelimiterspace} 2}$. And we have $\left\|g_2 \ensuremath\mathop{\scalebox{.65}{$\circ$}}\textbf{\emph{g}}_1\left(\textbf{\emph{x}}\right)-f_2\ensuremath\mathop{\scalebox{.65}{$\circ$}}\textbf{\emph{g}}_1\left(\textbf{\emph{x}}\right)\right\|<{\delta  \mathord{\left/ {\vphantom {\delta  4}} \right. \kern-\nulldelimiterspace} 2}$ according to Theorem \ref{theorem1}. Hence,
\begin{equation}
\left\|g_2\ensuremath\mathop{\scalebox{.65}{$\circ$}}\textbf{\emph{g}}_1\left(\textbf{\emph{x}}\right)- f_2\ensuremath\mathop{\scalebox{.65}{$\circ$}}\textbf{\emph{f}}_1 \left(\textbf{\emph{x}}\right) \right\| <{\delta  \mathord{\left/ {\vphantom {\delta  4}} \right. \kern-\nulldelimiterspace} 2} +{\delta  \mathord{\left/ {\vphantom {\delta  4}} \right. \kern-\nulldelimiterspace} 2}=\delta
\label{eq4}
\end{equation}
for every $\textbf{\emph{x}} \in E^n$. If we assume $M=3$, then
\begin{equation}\begin{aligned}
\left\| {G \left(\textbf{\emph{x}}\right) - F \left(\textbf{\emph{x}}\right)} \right\| &=\left\|g_3\ensuremath\mathop{\scalebox{.65}{$\circ$}}\textbf{\emph{g}}_2\ensuremath\mathop{\scalebox{.65}{$\circ$}}\textbf{\emph{g}}_1\left(\textbf{\emph{x}}\right)- f_3\ensuremath\mathop{\scalebox{.65}{$\circ$}}\textbf{\emph{f}}_2\ensuremath\mathop{\scalebox{.65}{$\circ$}}\textbf{\emph{f}}_1\left(\textbf{\emph{x}}\right) \right\| \\
&\le \left\|g_3\ensuremath\mathop{\scalebox{.65}{$\circ$}}\textbf{\emph{g}}_2\ensuremath\mathop{\scalebox{.65}{$\circ$}}\textbf{\emph{g}}_1\left(\textbf{\emph{x}}\right)- f_3\ensuremath\mathop{\scalebox{.65}{$\circ$}}\textbf{\emph{g}}_2\ensuremath\mathop{\scalebox{.65}{$\circ$}}\textbf{\emph{g}}_1\left(\textbf{\emph{x}}\right)\right\|\\
&+\left\|f_3\ensuremath\mathop{\scalebox{.65}{$\circ$}}\textbf{\emph{g}}_2\ensuremath\mathop{\scalebox{.65}{$\circ$}}\textbf{\emph{g}}_1\left(\textbf{\emph{x}}\right)- f_3\ensuremath\mathop{\scalebox{.65}{$\circ$}}\textbf{\emph{f}}_2\ensuremath\mathop{\scalebox{.65}{$\circ$}}\textbf{\emph{f}}_1\left(\textbf{\emph{x}}\right) \right\|
\end{aligned}
\label{eq5}
\end{equation}
Since $f_3$ is continuous in ${R^{{n \mathord{\left/{\vphantom {n {\prod\nolimits_{j = 1}^{2} {{k_j}} }}} \right. \kern-\nulldelimiterspace} {\prod\nolimits_{j = 1}^{2} {{k_j}} }}}}$, then according to Eq (\ref{eq4}), there exists $\eta >0$ such that $\left\|f_3\ensuremath\mathop{\scalebox{.65}{$\circ$}}\textbf{\emph{g}}_2\ensuremath\mathop{\scalebox{.65}{$\circ$}}\textbf{\emph{g}}_1\left(\textbf{\emph{x}}\right)- f_3\ensuremath\mathop{\scalebox{.65}{$\circ$}}\textbf{\emph{f}}_2\ensuremath\mathop{\scalebox{.65}{$\circ$}}\textbf{\emph{f}}_1\left(\textbf{\emph{x}}\right) \right\|<{\eta  \mathord{\left/{\vphantom {\eta  2}} \right.\kern-\nulldelimiterspace} 2}$. And according to Theorem \ref{theorem1}, it's easy to show that $\left\| g_3\ensuremath\mathop{\scalebox{.65}{$\circ$}}\textbf{\emph{g}}_2\ensuremath\mathop{\scalebox{.65}{$\circ$}}\textbf{\emph{g}}_1\left(\textbf{\emph{x}}\right)- f_3\ensuremath\mathop{\scalebox{.65}{$\circ$}}\textbf{\emph{g}}_2\ensuremath\mathop{\scalebox{.65}{$\circ$}}\textbf{\emph{g}}_1\left(\textbf{\emph{x}}\right)\right\|<{\eta  \mathord{\left/{\vphantom {\eta  2}} \right.\kern-\nulldelimiterspace} 2}$. Hence
 \begin{equation}
 \left\|g_3\ensuremath\mathop{\scalebox{.65}{$\circ$}}\textbf{\emph{g}}_2\ensuremath\mathop{\scalebox{.65}{$\circ$}}\textbf{\emph{g}}_1\left(\textbf{\emph{x}}\right)- f_3\ensuremath\mathop{\scalebox{.65}{$\circ$}}\textbf{\emph{f}}_2\ensuremath\mathop{\scalebox{.65}{$\circ$}}\textbf{\emph{f}}_1 \left(\textbf{\emph{x}}\right) \right\| <{\eta  \mathord{\left/ {\vphantom {\delta  4}} \right. \kern-\nulldelimiterspace} 2} +{\eta  \mathord{\left/ {\vphantom {\eta  4}} \right. \kern-\nulldelimiterspace} 2}=\eta
 \label{eq6}
 \end{equation}
 for every $\textbf{\emph{x}} \in E^n$. By following the line of proof of the above, we can finally prove that there exists $\varepsilon  > 0$ such that
\begin{equation}
\left\|g_{M}\ensuremath\mathop{\scalebox{.65}{$\circ$}} \textbf{\emph{g}}_{M-1} \ensuremath\mathop{\scalebox{.65}{$\circ$}} \cdots \ensuremath\mathop{\scalebox{.65}{$\circ$}}\textbf{\emph{g}}_1\left(\textbf{\emph{x}}\right)-f_{M}\ensuremath\mathop{\scalebox{.65}{$\circ$}} \textbf{\emph{f}}_{M-1} \ensuremath\mathop{\scalebox{.65}{$\circ$}} \cdots \ensuremath\mathop{\scalebox{.65}{$\circ$}}\textbf{\emph{f}}_1\left(\textbf{\emph{x}}\right)\right\| <\varepsilon
\label{eq7}
\end{equation}
for every $\textbf{\emph{x}} \in E^n$. Consequently, the non-overlapping CNN $G$ can approximate an arbitrary continuous function from $E^n$ to $\mathbb{R}$. This completes the proof. To make Theorem \ref{theorem2} clearer, we introduce Figure \ref{fig:2} which shows the structure of the non-overlapping CNN $G$.
\end{proof}

In the language of convolutional neural networks, non-overlapping CNNs with $M$ hidden layers and an unbiased convolutional layer behind each hidden layer are capable of approximating arbitrary continuous functions from $E^n$ to $R$, provided only that sufficiently many hidden neurons are available. Since the values of $n$, $M$ and $k_i$ are not determined, we can freely build the structures of non-overlapping CNNs under the condition of $n = \prod\nolimits_{i = 1}^M {{k_i}} $. This may make our theorems have much practical utility. In what follows, we briefly introduce the approximation capabilities of non-overlapping CNNs with multi-output neurons.

\newtheorem{theorem3}[theorem]{Theorem}
\begin{theorem3}\label{theorem4}
Let $G$ and $\textbf{{x}}$ be as defined in Theorem \emph{\ref{theorem2}} and $m \in \mathbb{N}$. Then the non-overlapping CNN $\textbf{{G}}$ with $m$ output neurons can approximate an arbitrary continuous function from $E^n$ to $\mathbb{R}^m$.
\end{theorem3}
\begin{proof}[\rm \textbf{Proof.}]
Since a multi-output network can be regarded as composed of multiple one-output networks, the non-overlapping CNN $\textbf{{G}}$ with $m$ output neurons is formulated as $\textbf{\emph{G}}\left( \textbf{\emph{x}} \right) = \left( {{G_1}\left( \textbf{\emph{x}} \right),{G_2}\left( \textbf{\emph{x}} \right), \cdots ,{G_m}\left( \textbf{\emph{x}} \right)} \right)$. For $i \ne j$ with $i,j = 1,2, \cdots ,m$, the difference between $G_i$ and $G_j$ is only the output neuron. It's clear that $\textbf{\emph{F}} \left( \textbf{\emph{x}} \right)= \left( {{F_1}\left( \textbf{\emph{x}} \right),{F_2}\left( \textbf{\emph{x}} \right), \cdots ,{F_m}\left( \textbf{\emph{x}} \right)} \right)$ is continuous in $E^n$ if and only if $F_1,F_2, \cdots, F_m$ are continuous in $E^n$ (see \cite{Rudin1964}, Chapter 4, pp.89-93.). Let $F_i$ be continuous in $E^n$, then according to Theorem \ref{theorem2}, there exists $\varepsilon  > 0$ such that
\begin{equation}\label{eq8}
\left\| {\textbf{\emph{G}}\left( \textbf{\emph{x}} \right) - \textbf{\emph{F}}\left( \textbf{\emph{x}} \right)} \right\| \le \sum\nolimits_{i = 1}^m {\left\| {{G_i}\left( \textbf{\emph{x}} \right) - {F_i}\left( \textbf{\emph{x}} \right)} \right\|}  < m \times \frac{\varepsilon }{m} = \varepsilon
\end{equation}
for $\left\| {{G_i}\left( \textbf{\emph{x}} \right) - {F_i}\left( \textbf{\emph{x}} \right)} \right\| < {{\varepsilon  \mathord{\left/ {\vphantom {\varepsilon  m}} \right. \kern-\nulldelimiterspace} m}}$. Therefore, the non-overlapping CNN $\textbf{\emph{G}}$ with $m$ output neurons can approximate an arbitrary continuous function from $E^n$ to $\mathbb{R}^m$.
\end{proof}

\section{Numerical experiments}
\label{section4}
The efficiency of our results is validated in the compressive sensing paradigm which is the problem of recovering a high dimensional and unknown signal $\textbf{\emph{y}} \in {R^m}$ from a low dimensional measurement $\textbf{\emph{x}} \in {R^n}$, $n<m$. The noisy measurement formula of compressive sensing is: $\textbf{\emph{x}} = A\textbf{\emph{y}} + \textbf{\emph{v}}$, where $A \in R^{n \times m}$ is the measurement matrix and $\textbf{\emph{v}} \in {R^n}$ is noise. In this paper, the measurement matrix $A$ is a random Gaussian matrix \emph{i.e.} ${A_{i,j}} \sim {\cal N}\left( {0,{1 \mathord{\left/ {\vphantom {1 n}} \right.\kern-\nulldelimiterspace} n}} \right)$, and the noise $\textbf{\emph{v}} \sim {\cal N}\left( 0,\delta \textbf{\emph{I}} \right)$, where $\delta$ represents the noise level. All non-overlapping CNN experiments are performed on Pytorch framework using a NVidia GeForce GTX 1660 Ti GPU.

\subsection{Non-overlapping CNN}
\begin{figure}[t]
\centering
\includegraphics[width=1\textwidth]{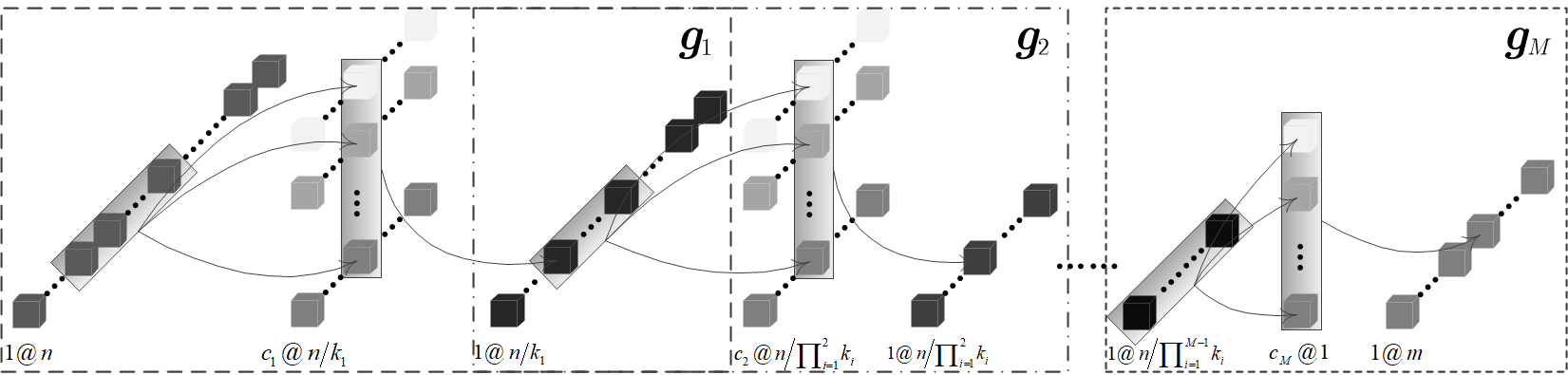}
\caption{Simplified illustration of the non-overlapping CNN}
\label{fig:3}       
\end{figure}

According to Theorem \ref{theorem4}, we give a simplified illustration of the non-overlap- ping CNN in Figure \ref{fig:3}. We observe that the non-overlapping CNN consists of $M$ networks $\textbf{\emph{g}}_i$ from Theorem \ref{theorem2}. In Figure \ref{fig:3}, $n$ represents the input dimension which matches the dimension of the measurement $\textbf{\emph{x}}$ and $m$ represents the output dimension which matches the dimension of the signal $\textbf{\emph{y}}$ to be recovered as well as $k_i$ and ${c_i}$ represent the length and number of convolution kernels in the network $\textbf{\emph{g}}_i$ respectively.

\begin{table}[b]
\centering
\caption{Comparison results of non-overlapping CNN with different loss under CelebA dataset. The bold numbers denote the best performance and the underlined numbers denote the second-best performance.}
\label{table:1}
\begin{tabular}{ccccc}
\hline
Criterion & L1     & MSE          & MSE+L1          &  $(1-\alpha)$MSE+$\alpha$L1 \\ \hline
RRE       & 0.1286 & {\ul 0.1274} & 0.1281          & \textbf{0.1273} \\
PSNR      & 22.595 & {\ul 22.665} & 22.633          & \textbf{22.677} \\
SSIM      & 0.7035 & 0.6982       & \textbf{0.7042} & {\ul 0.7038}    \\ \hline
\end{tabular}
\end{table}

To make all network parameters have better adaptability to the input measurement, a new loss function is proposed in this paper. The loss function is given by
\begin{equation}\label{eq9}
Loss = \frac{1 - \alpha }{N}\sum\limits_{i = 1}^N {{L_{MSE}}\left( {\textbf{\emph{G}}\left( {{\textbf{\emph{x}}^i}} \right),{\textbf{\emph{y}}^i}} \right)}  + \frac{\alpha }{N}\sum\limits_{i = 1}^N {{L_{L1}}\left( {G\left( {{\textbf{\emph{x}}^i}} \right),{\textbf{\emph{y}}^i}} \right)}
\end{equation}
where $\left\{ {{\textbf{\emph{x}}^i}} \right\}_{i = 1}^N$ and $\left\{ {{\textbf{\emph{y}}^i}} \right\}_{i = 1}^N$ are the training measurement set and the true signal set, respectively. $\alpha$ is the relaxation parameter. $L_{MSE}$ is the mean squared error (MSE) loss function which could reduce the error between the recovered signal $\textbf{\emph{G}}\left( {{\textbf{\emph{x}}^i}} \right)$ and the corresponding true signal $\textbf{\emph{y}}^i$, while it usually leads to the loss of information on signal details due to its oversmoothness. $L_{L1}$ is the $L1$ loss function which aims to make the element distribution of the recovered signal as close as possible to that of the corresponding true signal. In Table \ref{table:1}, we provide the performance of different loss function and the implementation details of all experiments are completely consistent. Obviously, the Introduction of $L1$ loss function alleviates the problem of oversmoothness. Compare with a simple combination of MSE and L1, the proposed loss function provides a trade-off performance.

\subsection{Datasets and implementation details}
We use two published datasets for our experiments.
\par \noindent \textbf{MNIST} dataset consists of $28 \times 28$ images of handwritten digits with 60,000 samples. We randomly select 55,000 images for training the non-overlapping CNN, the rest 5,000 images are used for evaluation. We set the input dimension $n=400$ and the noise level $\delta=0.01$. Here, the non-overlapping CNN consists of 2 networks $\textbf{\emph{g}}_1$ and $\textbf{\emph{g}}_2$, we set $k_1=2$ and $k_2=200$ as well as $c_1=256$ and $c_2=768$. The non-overlapping CNN is trained for 350 epochs using the AdamW optimizer with step-drop learning rate because it outperforms Adam optimizer with ${L_{\rm{2}}}$ regularization. The parameters used for AdamW optimizer follow the idea of the original paper: the initialized learning rate is setting to 0.001, ${\beta _{\rm{1}}}{\rm{ = 0}}{\rm{.9}}$, ${\beta _{\rm{2}}}{\rm{ = 0}}{\rm{.999}}$ and $\epsilon={\rm{1}}{{\rm{0}}^{{\rm{ - 8}}}}$. The learning rate is updated by multiplying the current learning rate by 0.7 after every 70 epochs. Other hyper-parameters are chose through many experiments. In particular, the weight decay factor is setting to 0.003, the relaxation parameter $\alpha=0.05$ and a mini-batch size of 500 is required to satisfy the requirement of the training set. For weights initialization, we use the Glorot initialization for each convolution kernel.
\par \noindent \textbf{CelebA} dataset consists of more than 200,000 number of face images. We select 120,000 high-quality images and crop them into $64\times 64$ greyscale images. Each pixel value was scaled to $\left[ { 0,1} \right]$. We randomly select 110,000 images to train the network and then use the rest 10,000 images to evaluate the trained model. Here, the input dimension $n$ is setting to 500 and the noise level $\delta$ is the same as the MNIST dataset. The non-overlapping CNN also consists of 2 networks $\textbf{\emph{g}}_1$ and $\textbf{\emph{g}}_2$, the length and number of convolution kernels are $k_1=1$ and $k_2=500$ as well as $c_1=512$ and $c_2=3072$. The non-overlapping CNN is trained using AdamW optimizer with the relaxation parameter $\alpha=0.15$, a initialized learning rate of 0.0003 and a mini-batch size of 300 for 300 epochs. The learning rate is updated by multiplying the current learning rate by 0.7 after every 50 epochs. Other hyper-parameters and weights initialization are the same as the MNIST dataset.

In this paper, we compare the non-overlapping CNN on MNIST and CelebA with algebraic reconstruction technique (ART) and Lasso for linear inverse problem, namely compressive sensing. To evaluate the performance of different algorithms, we use the relative restoration error (RRE), ${{\left\| {\textbf{\emph{G}}\left( \textbf{\emph{x}} \right) - \textbf{\emph{y}}} \right\|} \mathord{\left/ {\vphantom {{\left\| {\textbf{\emph{G}}\left( \textbf{\emph{x}} \right) - \textbf{\emph{y}}} \right\|} {\left\| \textbf{\emph{y}} \right\|}}} \right. \kern-\nulldelimiterspace} {\left\| \textbf{\emph{y}} \right\|}}$, peak signal-to-noise ratio (PSNR), $20{\log _{10}}\left( {{1 \mathord{\left/ {\vphantom {1 {\sqrt {{{\left\| {\textbf{\emph{G}}\left( \textbf{\emph{x}} \right) - \textbf{\emph{y}}} \right\|} \mathord{\left/{\vphantom {{\left\| {\textbf{\emph{G}}\left( \textbf{\emph{x}} \right) - \textbf{\emph{y}}} \right\|} m}} \right.\kern-\nulldelimiterspace} m}}}}} \right. \kern-\nulldelimiterspace} {\sqrt {{{\left\| {\textbf{\emph{G}}\left( \textbf{\emph{x}} \right) - \textbf{\emph{y}}} \right\|} \mathord{\left/{\vphantom {{\left\| {\textbf{\emph{G}}\left( \textbf{\emph{x}} \right) - \textbf{\emph{y}}} \right\|} m}} \right.\kern-\nulldelimiterspace} m}}}}} \right)$, and structural similarity index (SSIM).

\subsection{Evaluations}
In Table \ref{table:2}, we provide the average accuracy of different algorithms on MNIST and CelebA. non-overlapping CNN consistently outperforms ART and Lasso in terms of RRE, PSNR and SSIM.

\begin{table}[h]
\centering
\caption{Average comparison results of non-overlapping CNN with ART and Lasso. The bold numbers denote the best performance.}
\label{table:2}
\begin{tabular}{cccccc}
\hline
\multicolumn{2}{c}{Method}           & \multirow{2}{*}{ART} & \multirow{2}{*}{Lasso} & \multirow{2}{*}{Non-overlapping CNN} \\ \cline{1-2}
Dataset                 & Criterion  &                      &                        &                                      \\ \hline
\multirow{3}{*}{MNIST}  & RRE        & 0.7021               & 0.2808                 & \textbf{0.2608}                      \\
                        & PSNR       & 12.874                & 20.989                  & \textbf{21.683}                       \\
                        & SSIM       & 0.2199               & 0.7334                 & \textbf{0.8856}                      \\ \hline
\multirow{3}{*}{CelebA} & RRE        & 0.8413               & 0.1338                 & \textbf{0.1273}                      \\
                        & PSNR       & 6.013                 & 22.210                  & \textbf{22.677}                       \\
                        & SSIM       & 0.0357               & 0.6749                 & \textbf{0.7038}                      \\ \hline
\end{tabular}
\end{table}

\begin{figure*}[h]
  \includegraphics[width=1\textwidth]{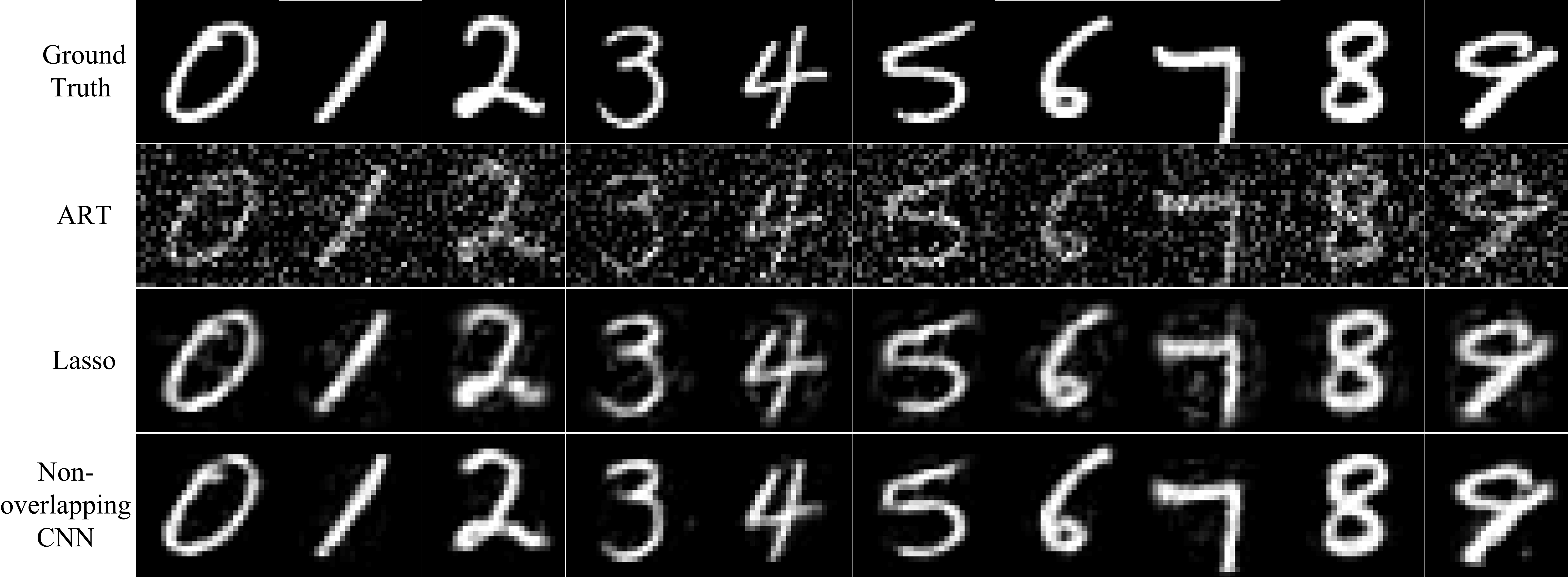}
\caption{Visual evaluation of different algorithms on MNIST.}
\label{fig:1}       
\end{figure*}

\begin{figure*}[h]
  \includegraphics[width=1\textwidth]{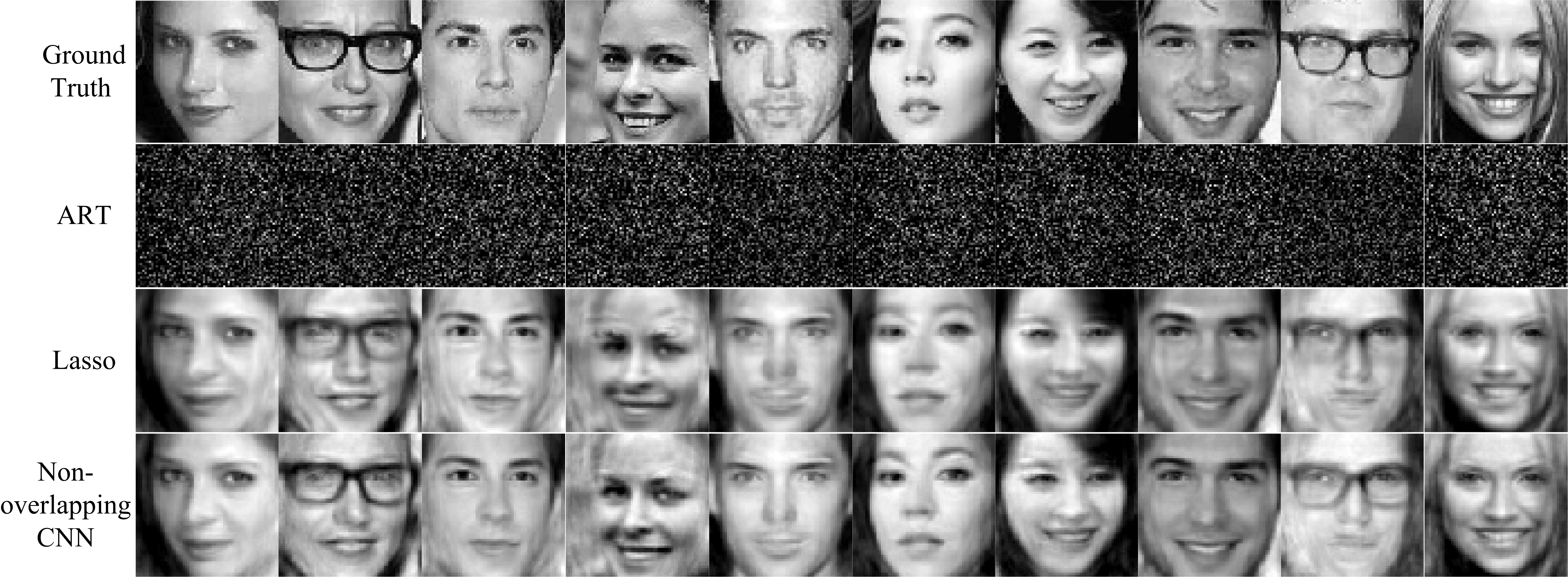}
\caption{Visual evaluation of different algorithms on CelebA.}
\label{fig:2}       
\end{figure*}

Moreover, we present some visual results in Figure \ref{fig:1} and \ref{fig:2} to compare the performance of non-overlapping CNN with the other two algorithms in terms of visual quality. Figure \ref{fig:1} and \ref{fig:2} show the reconstruction of ten test images from MNIST and eight test images from CelebA, respectively. From these figures, we can observe that non-overlapping CNN visually outperforms ART and Lasso. For the visual results in Figure \ref{fig:1}, the images reconstructed by non-overlapping CNN contain few artifacts and are clearer than Lasso and ART. And in Figure \ref{fig:2}, they are also superior to ART and Lasso as it is able to preserve detailed facial features such as glasses, eyes and mouth. In addition, although the measurements are severely deteriorated by noise, non-overlapping CNN is still able to reconstruct clearer images than Lasso which indicates non-overlapping CNNs are less sensitive to noise.

\section{Conclusion}
In this paper, a class of CNNs called non-overlapping CNNs are proposed to solve linear ill-posed inverse problems. We theoretically prove that non-overlapping CNNs with sigmoidal activation function can approximate arbitrary continuous function defined on compact input sets with any desired degree of accuracy. Evaluations elucidate the relative merits of non-overlapping CNNs in terms of performance and indicate that non-overlapping CNNs are less sensitive to noise. Our results are also suitable for overlapping CNNs. The operand calculation of the overlapping convolutional layer can be defined as follows:
\newtheorem{definition3}[theorem]{Definition}
\begin{definition3}\label{definition3}
Let $\textbf{{x}}$, $n$ and $k$ be as defined in Definition \emph{\ref{definition1}}. For $s \in \mathbb{N}$ with ${{\left( {n - k} \right)} \mathord{\left/ {\vphantom {{\left( {n - k} \right)} s}} \right. \kern-\nulldelimiterspace} s} \in \mathbb{N}$ and $s < k$, $\textbf{{x}}_j = \left( {{x_{js}},{x_{js + 1}}, \cdots ,{x_{js+k - 1}}} \right) \in E^k$ where $j = 0,1, \cdots ,{{\left( {n - k} \right)} \mathord{\left/{\vphantom {{\left( {n - k} \right)} s}} \right. \kern-\nulldelimiterspace} s}$. $A^k$ is the set of all affine functions from ${E^k}$ to $\mathbb{R}$, that is, the set of all functions of the form ${A\left( \textbf{{x}}_j \right)} = \textbf{{w}} \cdot {\textbf{{x}}_j} + b = \sum\nolimits_{i = js}^{k+js - 1} {{w_{i - js}}} {x_i} + b$ where $\textbf{{w}} \in {\mathbb{R}^k}$ and $b \in \mathbb{R}$. $\textbf{{A}} \left( \textbf{{x}} \right)=\left( A\left( {{\textbf{{x}}_0}} \right),A\left( {{\textbf{{x}}_1}} \right), \cdots ,A\left( {{\textbf{{x}}_{{\left( {n - k} \right)} \mathord{\left/ {\vphantom {{\left( {n - k} \right)} s}} \right.\kern-\nulldelimiterspace} s}}}\right) \right)^T$ is the output of the overlapping convolutional layer.
\end{definition3}
\par \noindent The proof of corresponding theorem is similar to that of non-overlapping CNNs. And we can freely build the structures of overlapping CNNs under the condition of $n - \sum\nolimits_{i = 1}^{\tilde M} {\left( {{k_i}\prod\nolimits_{j = 0}^{i - 1} {{s_j}}  - \prod\nolimits_{j = 1}^i {{s_j}} } \right) = \prod\nolimits_{j = 1}^{\tilde M} {{s_j}} }$. Observing that the length of hidden neurons in the overlapping CNN is larger than that in the non-overlapping CNN for the same task if the number of layers in the two networks is the same and the structure of the overlapping CNN is deeper than that of the non-overlapping CNN for the same task if the length of hidden neurons in the two networks is the same, since $s<k$ in the overlapping CNN and $s=k$ in the non-overlapping CNN. In other words, we need more data to train the overlapping CNN to avoid overfitting. An important area for further investigation is accurate assessment of the number of hidden neurons, investigation of this problem may be facilitated by consideration of Kolmogorov's representation theorem.


\end{document}